\documentclass[cameraready]{Interspeech}
\usepackage{xcolor}

\title{MEUSLI: a Multilingual Projector for LLM-based ASR and Beyond}

 \author[affiliation={1}] {Lorenzo} {Concina}
 \author[affiliation={1, 2}, orcid=0000-0002-8256-746X] {Seraphina} {Fong}
 \author[affiliation={1}] {Marco} {Matassoni}
 \author[affiliation={1}, orcid=0000-0003-4146-3071] {Alessio} {Brutti}

\address{
 $^1$ Center for Augmented Intelligence, Fondazione Bruno Kessler, Trento, Italy \\
 $^2$ Department of Information Engineering and Computer Science, University of Trento, Italy
}
\email{\{lconcina, mfong, brutti, matasso\}@fbk.eu}

\usepackage{comment}
\usepackage{multirow} 
\usepackage{pifont}

\begin{document}
\maketitle
\begin{abstract}
Lightweight projectors are an established way to connect pre‑trained speech encoders with large language models (LLMs), mapping acoustic features into token‑level embeddings for tasks like ASR and spoken question answering. Existing systems, however, typically only support a few languages and are often limited to English. We introduce MEUSLI, the first open‑science multilingual projector family that links a Whisper encoder with open‑source multilingual LLMs, enabling fully open‑source end‑to‑end ASR in 28 European languages. MEUSLI extends prior monolingual pipelines, delivering strong results across high‑ and low‑resource languages. Using proper continual leaning techniques, MEUSLI can be easily extended to other languages not seen in training. 

We further demonstrate that the MEUSLI projector can be leveraged beyond ASR, enabling multilingual speech translation and topic identification with only a few hours of task-specific supervision per language. 

Overall, MEUSLI provides a solid foundation for multilingual speech understanding tasks, supporting scalable and inclusive open-source SpeechLLMs.


\end{abstract}

\section{Introduction}
\label{sec:intro}
Large Language Models (LLMs) have transformed natural language processing, enabling systems that generalize across diverse text‑based tasks. Extending these capabilities to speech is essential for natural multimodal interaction, motivating the rise of Speech Language Models (SLMs), which allow LLMs to process audio directly~\cite{surveySLM}.
%
%
Traditional speech‑enabled LLMs rely on cascaded pipelines: automatic speech recognition (ASR), LLM processing, and optionally text-to-speech (TTS). Although effective, these architectures suffer from error propagation, loss of paralinguistic cues~\cite{audioVisualASR}, lack of shared context, and increased latency~\cite{recentAdvancesSLM}. These limitations have driven interest in unified SLMs that integrate speech and language modeling.

\begin{figure}[htbp]
  \centering
  
  \includegraphics[width=0.8\linewidth]{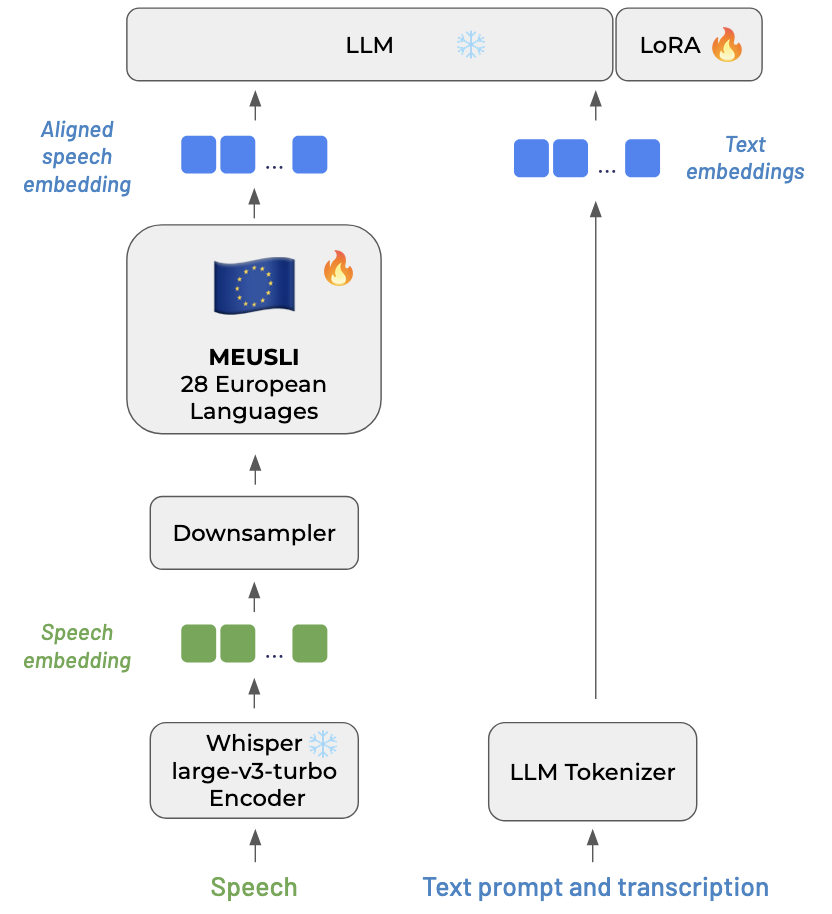}

  \caption{The proposed MEUSLI (Multilingual EU Speech LInear projector) training pipeline, following the SLAM-ASR architecture from~\cite{ma2024embarrassingly, fong25_interspeech}.}
  \label{fig:slm}
\end{figure}

Recent multimodal systems such as Qwen‑Audio~\cite{chu2023qwen,chu2024qwen2} and SALMONN~\cite{tang2023salmonn} show the potential of directly connecting speech encoders to LLMs. However, they typically support only high‑resource languages, offer limited adaptability, or rely on proprietary data. SLMs based on SLAM‑ASR~\cite{ma2024embarrassingly} mitigate these issues by introducing trainable projectors that map acoustic representations from a pre‑trained encoder into the LLM embedding space (Figure~\ref{fig:slm}).

The speech encoder converts audio into high‑dimensional features, which the projector maps into the LLM’s embedding space. Projectors range from simple linear layers to more complex architectures~\cite{Yu2023ConnectingSE,Geng2024UnveilingTP}, often with temporal downsampling to reduce LLM costs. During training, the projector learns to align acoustic and textual representations within a unified generative framework, enabling end‑to‑end speech tasks such as ASR, speech translation, and spoken language understanding~\cite{Yang2024CTCAssistedLC,Shi2024AdvancingMA,Ma2025SpeechRM,geng2025osum}

Despite growing interest in lightweight SLMs~\cite{verdini25_interspeech, SLAM-ASR-eval}  and prompting strategies~\cite{Ma2024EffectiveTA, burdisso_promptor}, current approaches remain largely English-centric and limited in multilingual scope. Key gaps persist in low-resource language support, multilingual adaptation strategies, systematic analysis of training data composition or specialized encoders, and reproducible open research practices.

\subsection{Related Work}
Recent work has begun to address these gaps, including adaptation for low-resource Indic languages \cite{mittal24_salsa_interspeech,mittal25_skip_salsa_interspeech}, robustness to pathological speech \cite{zhang25t_atypical_asr_llm_interspeech}, and multilingual ASR in eight languages \cite{fathullah2024prompting}. Moreover, \cite{denisov-vu-2024-multilingllm} proposed a multilingual SLM trained on 1900 hours of speech from 139 languages. Although this demonstrates strong multilingual coverage, adaptability to low-resource languages is underexplored. High-resource pre-training with minimal data fine-tuning has the potential to help mitigate this gap \cite{fong25_interspeech}, and recent work has shown that grouping languages by family for shared projectors can improve multilingual robustness and cross-language transfer \cite{zhang2026languagefamily}. However, systematic investigation in multilingual settings at larger scales remains limited. Finally, although BLOOMZMMS~\cite{denisov-vu-2024-multilingllm} supports a large number of languages, the impact of the training material or the combination of specialized encoders is not explored. A further limitation of models currently available in the literature is the lack of openness of the data used for training and the associated licensing of the model, which hinders reproducibility and transparency. 
Table~\ref{tab:sota} summarizes current models and their limitations in terms of number of languages, inclusion of low-resource ones, tasks, and their openness. 

\begin{table}[htbp]
    \caption{Survey of recent open source SLMs.}
    \label{tab:sota}
    \centering
    \begin{tabular}{|l|l|c|c|c|}
    \hline
   
    \multirow{2}{*}{Model}      &  \multirow{2}{*}{Lang.}        & Low   & \multirow{2}{*}{Task} & \multirow{2}{*}{Data} \\        
                                & & Res & & \\ 
    \hline
       SLAM-ASR~\cite{ma2024embarrassingly,SLAM-ASR-eval} & En & \ding{55} & multi & open \\
       SpeechLMM~\cite{verdini25_interspeech} & 5 & \ding{55} & multi  & open\\
       SALMONN~\cite{tang2023salmonn} & En & \ding{55} & multi & open-weights \\
       Qwen-Audio~\cite{chu2024qwen2} & 8 & \ding{55} &  multi& open-weights\\
       SALSA\cite{mittal24_salsa_interspeech,mittal25_skip_salsa_interspeech}  & 8 & \ding{51} & ASR & open\\
       \cite{fathullah2024prompting} & 8 & \ding{55} & ASR & open-weights\\
       BLOOMZMMS~\cite{denisov-vu-2024-multilingllm} & 102 & \ding{51} & multi & open \\
        \hline
        {\bf MEUSLI} & 28+ & \ding{51} & ASR & open\\
        \hline
    \end{tabular}

\end{table}

\section{MEUSLI: a Multilingual Projector for 28 EU Languages (and Beyond)}
In this work, we address these gaps by introducing MEUSLI, the first fully open multilingual projector for LLM-based ASR supporting 28 European languages. MEUSLI extends the SLAM‑ASR architecture to a large multilingual setting by connecting a pre-trained Whisper encoder to open-source pre-trained multilingual LLMs via a lightweight linear projector. The resulting system enables efficient end‑to‑end ASR without modifying either the speech encoder or the LLM. Beyond multilingual ASR, we show that MEUSLI serves as a strong initialization point for fine-tuning on low-resource languages and can be used to bootstrap or extend ASR capabilities even for languages not seen during training.


While the SLAM architecture has been explored across multiple tasks, even in multi-task settings and for a limited number of languages, the present work focuses on multilingualism with ASR as a case study. This choice is motivated by the broad availability of ASR data across diverse languages, even if in small quantities for low-resource languages, and the existence of well-established evaluation benchmarks.  We show that, while trained for ASR, MEUSLI can be used as starting point to extend its capabilities to other speech processing tasks, in particular those for which limited data are available. Our results highlight the scalability, openness, and practicality of the approach, laying the groundwork for inclusive, LLM-powered speech technologies.

We build on our previous findings in \cite{fong25_interspeech}, which examined the capabilities of a SLAM-ASR approach in low-resource scenarios. We observed that approximately 100-200 hours of training data are needed to match Whisper-only performance. However, multilingual high-resource pretraining can greatly improve results in data-scarce conditions when fine-tuning with as little as 10 hours of data. 
Therefore, our goal in this paper is to extend those findings to a highly multilingual setting, obtaining an SLM that (i) supports most European (EU) languages, (ii) improves performance on low-resource languages, (iii) allows bootstrapping new ones and (iv) serves as starting point for other speech processing tasks. Overall, this paper  presents the following contributions: 
\begin{enumerate}
    \item {\bf MEUSLI} (Multilingual EU Speech language model LInear projector) extends the SLAM-ASR framework and related speech-based LLM paradigms from a limited set of languages to {\bf 28 European languages}. We release 
     a family of projectors on HuggingFace that support 28 EU Languages and are trained with several open source LLMs. \url{https://huggingface.co/collections/SpeechTek/meultilingual-speechllm-projectors} 
    \item We provide details of model configurations to facilitate successful fine-tuning within the SLAM-ASR framework.
    
    \item Unlike recent work on SpeechLLMs, we consider low-resource languages and demonstrate that MEUSLI can be effectively fine-tuned to support them, enabling new languages to be bootstrapped from only limited amounts of speech data. Moreover, we show that new languages can be added using state-of-the-art continual learning methodologies. 
    \item We show experimentally that, starting from MEUSLI, multilingual multitask projectors can be obtained with only a small amount of adaptation data, extending support beyond ASR to speech translation and topic identification. 
\end{enumerate}

To conclude, our work lays the foundation for scalable, open and inclusive speech-to-text systems powered by LLMs.

\subsection{The MEUSLI Model}
\label{sec:system}
In this work, we use the SLAM-ASR framework~\cite{ma2024embarrassingly} as the basis for our multilingual SLMs. 
We adopt \emph{Whisper-large-v3-turbo} as the encoder and 3 backend LLMs: \emph{EuroLLM~1.7B-Instruct}, \emph{EuroLLM~9B}~\cite{EUROLLM}, and \emph{Apertus-8B}~\cite{swissai2025apertus}. Following the SLAM-ASR methodology, both the encoder and the LLM are kept frozen, while only the linear projector is trained. 
%
While the projector design can range from simple linear layers to more elaborate architectures~\cite{surveySLM}, in this study we employ a linear projector, as it consistently resulted in the best ASR performance~\cite{Kumar-2025, concina25_mlcslm}. In our implementation, the projector consists of a single hidden layer with a ReLU activation, followed by a regression layer, comprising 17.31M trainable parameters. To further enhance performance, Low-Rank Adaptation (LoRA)~\cite{hu2022lowrank} was applied to the LLM. The best LoRA configuration was achieved using rank $r=8$ and scaling factor $\alpha=32$, introducing an additional 1.38M tunable parameters.

\subsection{Training Configuration}
\label{ssec:exp_setup_train_config}

Input audio was converted into mel spectrograms with 128 frequency bins. The output speech embeddings from the frozen speech encoder are downsampled by a factor of \textit{k} = 5 to mitigate the length mismatch between speech and text representations. The models were trained using Adam optimizer with an initial learning rate of $1 \times 10^{-4}$, a linear warmup of 1{,},000 steps and a subsequent scheduled learning rate decay. Training was carried out for 3 epochs with a batch size of 8 and a validation batch size of 2, balancing GPU memory constraints with training stability. The model was optimized using cross-entropy loss on the LLM outputs. During training, projected speech embeddings were concatenated with a text prompt (e.g., \textit{``Transcribe speech to text"}) and passed to the LLM.  Training was performed on one NVIDIA Ada Lovelace L40S GPU. This configuration served as the foundation for our multilingual experiments, providing stable training and competitive recognition accuracy across multiple European languages.

\subsection{Datasets}
\label{ssec:exp_setup_datasets}

We train the model using three widely used open speech datasets: Common Voice 17.0 (CV)~\cite{commonvoice:2020}, FLEURS (FL)~\cite{fleurs2022arxiv}, and VoxPopuli~\cite{wang-etal-2021-voxpopuli}. From these sources, we collected data spanning 28 European languages for a total of 7622 hours: English, French, German, Italian, Spanish, Portuguese, Dutch, Polish, Hungarian, Czech, Romanian, Bulgarian, Slovak, Slovenian, Serbian, Greek, Danish, Swedish, Finnish, Latvian, Lithuanian, Estonian, Welsh, Maltese, Breton, Irish, Galician, and Basque. To address the imbalanced data distribution and ensure computational feasibility, we capped the number of audio samples at 100K per language per dataset. This strategy reduced data skew and made training more efficient. 

Previous work~\cite{SLAM-ASR-eval,fong25_interspeech} has shown that SLMs can be sensitive to domain shifts and variations in speech conditions, such as noise, accents, or other perturbations. Therefore, we adopted an iterative training and evaluation approach which involved progressively adding new datasets, retraining the model, and measuring its performance to enhance MEUSLI's robustness by exposing it to a wider variety of audio samples. 
%
%
Furthermore, the model was evaluated not only on the test splits of the training datasets (in-domain), but also on out-of-domain data from the INTERSPEECH 2025 MLC-SLM Challenge\footnote{\url{https://www.nexdata.ai/competition/mlc-slm}}. Adding training data consistently improved the Word Error Rate (WER) across both in-domain and out-of-domain evaluations, validating our iterative multilingual training approach.

\section{Results and Discussion}
\label{sec:results_discussion}

We first evaluated the performance of the multilingual projector across the 28 European languages described in Section~\ref{ssec:exp_setup_datasets}. Table~\ref{tab:wer_multilingual} reports the WER obtained on the test splits of Common Voice (CV), FLEURS (FL), and the 2025 MLC-SLM Challenge (MLC). We compare the results obtained with the 3 LLMs against  \emph{Whisper-large-v3-turbo}. While the comparison against Whisper is not fully fair in terms of training data and resources, we keep it as a reference point for the performance in different languages.

Several trends emerge from the results. 
High-resource languages (top of Table~\ref{tab:wer_multilingual}) achieve strong transcription accuracy while low-resource languages (bottom of Table~\ref{tab:wer_multilingual}) are characterized by very high WERs. Since the training material is balanced across languages, these gaps largely reflect the accuracy of the speech encoder and of the LLM on specific languages (see Sec.~\ref{ssec:results_low_res_lang_ft}). 
This is also evident as the larger LLMs show better performance than EuroLLM-1.7B, with Apertus-8B slightly better on average. Concerning the comparison with Whisper, the LLM-based method using the larger models generally performs better, with major gains in particular for the low-resourced languages even when they are not originally covered by the pre-trained LLM. Note, however, that Whisper seems to be more effective on the MLC out-of-domain data, a behavior probably related to the larger amount of training data.

Overall, these results show that MEUSLI delivers strong performance across 28 languages: WERs are low for high-resource languages, and consistent improvements are observed for medium-resource ones. Although very low-resource languages such as Breton, Irish, and Maltese remain challenging, the model still demonstrates clear gains, confirming the scalability and robustness of the approach across diverse linguistic conditions. These results motivate further experiments on fine-tuning for low-resource languages and bootstrapping unseen ones, as discussed in Sec.~\ref{ssec:results_low_res_lang_ft} and~\ref{ssec:results_bootstrap}.

\begin{table*}[tb]
  \caption{WER on the 28 languages using 3 LLM-based models (EuroLLM-1.7B, EuroLLM-9B, Apertus-8B) and Whisper. While Apertus-8B supports all these languages, EuroLLM models don't cover those indicated with *.}
  \label{tab:wer_multilingual}
  \small
  \centering
  \begin{tabular}{l *{12}{c}}
    \toprule
    \textbf{Language}
    & \multicolumn{3}{c}{\textbf{Whisper L3-Turbo}}
    & \multicolumn{3}{c}{\textbf{EuroLLM-1.7B}}
    & \multicolumn{3}{c}{\textbf{EuroLLM-9B}}
    & \multicolumn{3}{c}{\textbf{Apertus-8B}} \\

    \cmidrule(lr){2-4} \cmidrule(lr){5-7}
    \cmidrule(lr){8-10} \cmidrule(lr){11-13}

    & CV & FL & MLC
    & CV & FL & MLC
    & CV & FL & MLC
    & CV & FL & MLC \\
    \midrule

    Spanish     & 7.60 & 3.05 & 15.70 & 5.22 & 4.09 & 21.86 & 3.65 & 3.53 & 24.51 & 4.49 & 3.76 & 25.22 \\
    German      & 8.85 & 4.71 & 20.77 & 7.11 & 7.79 & 32.77 & 5.22 & 5.84 & 38.50 & 5.51 & 5.37 & 34.01 \\
    Dutch       & 7.96 & 5.76 & --    & 6.83 & 8.65 & --    & 5.00 & 6.09 & --    & 5.45 & 6.32 & --    \\
    Portuguese  & 10.49& 3.95 & 34.92 & 9.39 & 4.86 & 51.75 & 7.23 & 3.83 & 45.32 & 8.75 & 4.47 & 48.41 \\
    Galician*   & 25.07& 12.54& --    & 12.70& 9.98 & --    & 9.01 & 6.81 & --    & 11.27& 8.43 & --    \\
    English     & 16.75& 4.91 & 8.14  & 12.94& 6.34 & 46.56 & 10.51& 4.68 & 18.34 & 11.23& 5.20 & 15.71 \\
    Polish      & 11.91& 5.41 & --    & 14.19& 8.68 & --    & 11.00& 6.06 & --    & 12.03& 6.21 & --    \\
    Czech       & 12.77& 11.22& --    & 11.16& 11.32& --    & 8.05 & 8.57 & --    & 9.25 & 5.58 & --    \\
    French      & 15.07& 5.47 & 32.74 & 11.24& 7.83 & 42.05 & 9.43 & 5.84 & 53.44 & 9.85 & 6.33 & 50.11 \\
    Hungarian   & 16.82& 14.32& --    & 14.59& 16.87& --    & 10.08& 12.72& --    & 14.86& 15.85& --    \\
    Italian     & 7.70 & 2.89 & 29.24 & 6.01 & 3.32 & 36.13 & 4.45 & 2.70 & 33.80 & 5.03 & 2.78 & 34.22 \\
    Swedish     & 20.95& 8.57 & --    & 15.99& 10.94& --    & 12.08& 8.09 & --    & 12.89& 8.58 & --    \\
    Romanian    & 20.62& 9.10 & --    & 17.39& 9.65 & --    & 14.42& 7.30 & --    & 15.31& 7.92 & --    \\
    Danish      & 18.15& 14.60& --    & 18.81& 14.65& --    & 14.41& 10.76& --    & 15.46& 10.43& --    \\
    Basque*     & 42.74& --   & --    & 19.96& --   & --    & 16.11& --   & --    & 12.14& --   & --    \\
    Bulgarian   & 19.88& 13.03& --    & 24.26& 15.20& --    & 18.63& 10.81& --    & 18.73& 10.37& --    \\
    Finnish     & 18.37& 8.42 & --    & 22.61& 15.29& --    & 17.94& 11.26& --    & 16.83& 11.06& --    \\
    Latvian     & 32.75& 18.75& --    & 27.12& 17.23& --    & 20.32& 11.92& --    & 19.51& 11.25& --    \\
    Lithuanian  & 36.07& 24.01& --    & 28.27& 24.30& --    & 18.60& 17.25& --    & 20.02& 17.06& --    \\
    Greek       & 26.39& 12.73& --    & 30.06& 18.35& --    & 24.24& 13.76& --    & 23.53& 14.22& --    \\
    Slovak      & 70.65& 15.40& --    & 35.84& 9.71 & --    & 25.97& 6.52 & --    & 29.98& 6.70 & --    \\
    Slovenian   & 39.20& 20.03& --    & 34.72& 19.41& --    & 28.17& 14.71& --    & 30.43& 13.73& --    \\
    Estonian    & 37.68& 19.00& --    & 37.19& 19.83& --    & 27.97& 15.40& --    & 28.76& 13.75& --    \\
    Welsh*      & 66.05& 34.47& --    & 50.40& 39.96& --    & 39.93& 33.41& --    & 39.25& 30.05& --    \\
    Serbian*    & 99.34& 22.78& --    & 56.49& 27.60& --    & 45.18& 23.74& --    & 35.29& 23.65& --    \\
    Maltese     & 98.39& 77.49& --    & 58.84& 44.89& --    & 48.56& 35.42& --    & 48.68& 39.87& --    \\
    Breton*     & 154.33& --  & --    & 95.68& --   & --    & 80.77& --   & --    & 73.40& --   & --    \\
    Irish       & 216.52& 609.95& --  & 82.23& 88.06& --     & 83.15& 84.51& --    & 88.55& 88.99& --    \\
    \bottomrule
  \end{tabular}
\end{table*}

\subsection{Further Finetuning on Low-resource Languages}
\label{ssec:results_low_res_lang_ft}
\begin{figure}[t]
  \centering
  
  \includegraphics[width=0.9\linewidth]{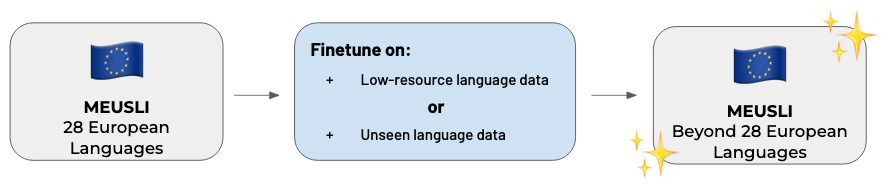}
  \caption{Illustration of the MEUSLI fine-tuning and bootstrapping pipeline for expanding language coverage.}
  \label{fig:slm_boostrap}
\end{figure}

Although the proposed pipeline demonstrates broad linguistic coverage, its performance is strongly dependent on the quantity and quality of available data per language, as well as the quality of the speech encoder and the LLM. This dependency extends beyond the projector training to encompass the datasets used to train both the speech encoder and the LLM. 
Monolingual systems tend to outperform their multilingual counterparts, but training them from scratch is rarely feasible for low-resource languages~\cite{fong25_interspeech}. MEUSLI provides an effective starting point to improve the performance of monolingual systems as shown in Figure~\ref{fig:slm_boostrap}. We illustrate this using Breton (BR) and Maltese (MT) as study case using \emph{EuroLLM~1.7B-Instruct}, which showed high error rates in the multilingual setting (Table \ref{tab:wer_multilingual}).

We fine-tuned the multilingual projector to a single language using, respectively: $\sim$18 hours of MT data from CV 20.0, FL, and MASRI-Headset v2 ~\cite{Williams2023TheAO} and $\sim$2.5 hours of BR data from CV 20.0. Training used the same setup as 
in Section \ref{ssec:exp_setup_train_config} with a $1 \times 10^{-4}$ learning rate, batch size 4, and 10 epochs. 
To test if language-specific features help, we also replaced the vanilla \textit{Whisper-large-v3-turbo} encoder with pre-fine-tuned and better performing versions for MT\footnote{\url{https://huggingface.co/carlosdanielhernandezmena/whisper-largev2-maltese-8k-steps-64h}} and BR\footnote{\url{https://huggingface.co/Bretagne/whisper-large-v3-turbo-audio_breton-transcription_breton}}. Table~\ref{tab:finetuning} presents a comparative analysis between: (i) fine-tuning our pretrained multilingual projector (MEUSLI), and (ii) training a monolingual system from scratch. The results indicate that initializing with our multilingual model results in substantial performance improvements for both languages.
Moreover, MEUSLI remains beneficial even when paired with a specialized monolingual encoder. In the case of BR, the pairing outperforms systems that rely on training a dedicated monolingual projector from scratch. It, moreover, remains competitive despite BR being unsupported by EuroLLM, highlighting the pipeline's robustness in low-resource settings.



\begin{table}[t]
    \caption{{\bf Low-resource}: WER when fine-tuning on Breton
(BR) and Maltese (MT). Training data: CV for BR; CV, FL, and MASRI-Headsetv2 for MT. ``Mono" refers to a monolingual projector trained on the aforementioned data, ``MEUSLI" is our projector, ``→” indicates MEUSLI fine-tuned. Encoder `Adapt' means a pre-fine-tuned and better performing Whisper encoder was used for either MT and BR. The LLM is \emph{EuroLLM~1.7B-Instruct}.}
    \label{tab:finetuning}
    \resizebox{\columnwidth}{!}{
    \centering
    \begin{tabular}{c|c|c|c|c|c}
        Lang. & Train. & Encoder & Mono & MEUSLI & MEUSLI→ \\
        \hline
        \multirow{2}{*}{Breton} & \multirow{2}{*}{2.5h} & Large-V3 & 84.3\% & 95.7\% & {\bf 79.7\%} \\
        & & Adapt & 27.7\% & - & {\bf 24.5\%}\\
        \hline
        \multirow{2}{*}{Maltese} & \multirow{2}{*}{18h} & Large-V3 & 45.8\% & 58.8\% & {\bf 39.4\%} \\
        & & Adapt  & \textbf{18.0\% }& - & 23.7\% \\        
    \end{tabular}
    }
\vspace{-0.3cm}
\end{table}

\begin{table}[t]
    \centering
    \caption{{\bf Beyond 28}: WER for the bootstrapping experiment on Ukranian and Albanian. The training data comes from CV. ``Mono" refers to a monoligual projector trained on the CV data, ``→" indicates MEUSLI fine-tuned on CV data. Note that both languages are not covered by EuroLLM.}
    \label{tab:bootstrapping}
    \begin{tabular}{c|c|c|c|c}
        Lang. & Train.  & Mono & MEUSLI & MEUSLI→ \\
        \hline
        Ukranian & 30h   & 20.4\% & 107\% & {\bf 16.3\%} \\ \hline
        Albanian & 46min  & 389\% & 166\% & {\bf 75.6\%}\\
    \end{tabular}
\end{table}

\subsection{Beyond 28: Bootstrapping for a New Language}
\label{ssec:results_bootstrap}
Building on the results obtained for BR and MT, we further explored the capacity of the multilingual projector to bootstrap ASR for languages not seen during training. We specifically investigated Ukrainian and Albanian, two low-resource European languages {\bf excluded from both the speech encoder and EuroLLM} pretraining corpora. As in Section~\ref{ssec:results_low_res_lang_ft}, we compared (i) fine-tuning MEUSLI on each language, and (ii) training a monolingual projector from scratch. For Ukrainian (30 hours of data), MEUSLI fine-tuning reached 16.34\% WER on the CV test set, against 20.43\% for the monolingual system and over 100\% for the unadapted multilingual baseline (Table~\ref{tab:bootstrapping}). For Albanian, with only 46 minutes of data, fine-tuning still achieved 75.61\% WER after 5 epochs, whereas a projector trained from scratch failed to converge (389.12\% WER). These results confirm that multilingual pretraining enables effective knowledge transfer: even with minimal data, MEUSLI bootstraps a functional ASR system for an unseen language, making the approach scalable to new low-resource languages.

\subsection{Mitigating Catastrophic Forgetting via Data Replay}
\label{ssec:results_cl}
Bootstrapping a new language comes at the cost of {\bf catastrophic forgetting}: the performance on the original 28 languages collapses as the projector overfits the new linguistic distribution. The effect is severe: after naively fine-tuning MEUSLI on Ukrainian, most base languages become untranscribable, and the model hallucinates Ukrainian tokens. This is shown in the first two rows of Table~\ref{tab:replay} which compare the original MEUSLI performance on a subset of languages with those obtained after finetuning on Ukrainian. Note, for example, that Spanish degrades from 4.09\% to 90.34\% WER and Polish from 8.68\% to 100.6\%.

A dedicated continual-learning study built on MEUSLI~\cite{Meusli-cl} shows that this trade-off can be resolved with rehearsal-based strategies (data replay)~\cite{rehearsalBased}. Interleaving a small buffer of samples from the original languages while learning the new one anchors the shared multilingual embedding space and prevents its collapse. Replaying as few as 1{,}000 samples per language restores the base languages to near-original WER while still acquiring Ukrainian at 17.56\% WER. This is in line with the 16.34\% WER obtained with naive fine-tuning, but without sacrificing prior knowledge (Table~\ref{tab:replay}). 

\begin{table}[t]
    \centering
    \caption{{\bf Data replay on Ukrainian}: WER (\%) for base MEUSLI (no adaptation), naive fine-tuning on Ukrainian (MEUSLI$\rightarrow$UK), and fine-tuning with a 1{,}000-sample-per-language replay buffer (+replay). Base languages are scored on FLEURS, Ukrainian (UK) on Common Voice.}
    \label{tab:replay}
    \resizebox{\columnwidth}{!}{
    \begin{tabular}{l|ccccccccc|c}
        & ES & DE & FR & IT & EN & DA & PL & CS & UK & Avg \\
        \hline
        MEUSLI                & 4.09 & 7.79 & 7.83 & 3.32 & 6.34 & 14.65 & 8.68  & 11.32 & - & 8.01 \\
        MEUSLI$\rightarrow$UK & 90.34 & 15.50 & 68.90 & 39.28 & 9.02 & 40.02 & 100.6 & 108.0 & 16.34 & 54.22 \\
        \quad +replay         & 4.82 & 9.54 & 7.96 & 5.94 & 5.20 & 18.67 & 10.95 & 15.21 & 17.56 & {\bf 10.65} \\
    \end{tabular}
    }
\end{table}

\section{Towards Multiple Tasks}
While MEUSLI demonstrates effective multilingual adaptation for low-resource ASR (Sections \ref{ssec:results_low_res_lang_ft} and \ref{ssec:results_bootstrap}), speech systems often need to support multiple downstream tasks beyond transcription. Recent work has explored multilingual or multitask SpeechLLMs \cite{mohapatra2026speechmapper,zufle-niehues-2025-contrastive-multitask,lee2025naver,wang25m_usam_interspeech,kim2025tesu,mundnich2025zero,janeiro-etal-2025-mixtureoflangs}, but multilingual multitask adaptation under data-scarce settings remains relatively unexplored. We therefore extend MEUSLI beyond ASR to show that leveraging highly multilingual projector pretraining can facilitate transfer to additional speech understanding tasks. In addition to ASR, we add speech translation and topic identification using less than 5 hours of labeled data per language-task pair. Further methodological details and experimental results are provided in our Interspeech 2026 paper, which will be made publicly available upon publication.

We evaluate two downstream speech tasks in addition to the ASR task:
\begin{itemize}
    \item \textbf{Speech Translation (ST).} The model generates English translations directly from speech input ($X \rightarrow EN$). Model translation quality is evaluated using BLEU \cite{post-2018-bleu}. 
    \item \textbf{Topic Identification (TID).} The model predicts a topic label from speech input. Performance is evaluated using accuracy and macro F1, with the latter metric accounting for class imbalance.

\end{itemize}

\subsection{Dataset}
We use the Italian, Spanish, Galician, Czech, and Finnish subsets of the SIB-Fleurs dataset \cite{schmidt2025fleursslumassivelymultilingualbenchmark,adelani2023sib200}. SIB-Fleurs contains parallel annotations for ASR, speech translation (ST), and topic identification (TID) on the same speech segments. This allows for multitask learning from identical audio inputs. Each language subset provides approximately 3--5 hours of labeled training data per task.

The selected languages span three language families (Romance, Slavic, and Uralic) to allow us to investigate transfer across varying degrees of linguistic similarity under low-resource conditions. For ST, we use the \textit{SeamlessM4T}-derived English translations provided by SIB-Fleurs. For TID, each utterance has a label that is one of seven topics: science/technology, travel, politics, sports, health, entertainment, and geography.

\subsection{Model Architecture and Training Configuration}
\label{ssec:multitask_training_method}
To extend MEUSLI beyond ASR, we employ the multitask training recipe provided by the SLAM-LLM repository\footnote{\url{https://github.com/X-LANCE/SLAM-LLM/blob/main/examples/aispeech_asr}} while keeping the architecture and optimization strategy unchanged. We utilize \textit{EuroLLM-1.7B-Instruct} as the LLM.

Training uses dynamic frame batching with \textit{train\_max\_frame\_length} = 1500 and \textit{eval\_max\_frame\_length} = 3000. We optimize the projector using cross-entropy loss over the target token sequence for all tasks, training for up to 10 epochs with early stopping based on validation loss. Beam search with beam size of 4 is used during inference.

The task-specific prompts used during training and inference include: ASR (\textit{``Transcribe speech to text."}), ST \cite{mundnich2025zero} (\textit{``Perform speech translation into English using the preceding audio:"}), and TID (`\textit{`Classify the topic of the spoken utterance into one of the following labels: science/technology, travel, politics, sports, health, entertainment, geography."}).

\subsection{Experimental Procedure}

As a baseline, we first train multitask linear projectors from-scratch: models are jointly optimized for ASR, ST, and TID on SIB-Fleurs using all available labeled data (less than five hours per language and task). Both multilingual (a single model trained jointly on all five languages) and monolingual (one model per language) training settings are considered. To then evaluate how MEUSLI extends beyond ASR, we finetune it on SIB-Fleurs under the same experimental settings. 

Comparing these four settings (from-scratch vs.\ MEUSLI initialization and monolingual vs.\ multilingual training) allows us to assess both the benefit of multilingual supervision and the transferability of MEUSLI multilingual representations to downstream tasks beyond ASR in data-scarce conditions.

\subsection{Results}

\begin{table*}[ht!]
\centering
\caption{Transferability of the pretrained MEUSLI projector beyond ASR. Comparison of multitask training from-scratch and MEUSLI-bootstrapped finetuning under multilingual (Multi) and monolingual (Mono) settings on SIB-Fleurs. SIB-Fleurs training data amounts per language/task: IT = $\sim$4.5hr, ES = $\sim$3.5hr, GL = $\sim$3.5hr, CS = $\sim$4hr, FIN = $\sim$4.5hr.}
\label{tab:meusli_multitask_scratch_vs_bootstrap}
\resizebox{\textwidth}{!}{%
\begin{tabular}{l cccc cccc cccc cccc}
\hline
& \multicolumn{4}{c}{\textbf{ASR}}
& \multicolumn{4}{c}{\textbf{ST}}
& \multicolumn{8}{c}{\textbf{Topic ID}} \\
\textbf{Language}
& \multicolumn{4}{c}{$\downarrow$ WER (\%)}
& \multicolumn{4}{c}{$\uparrow$ BLEU (\%)}
& \multicolumn{4}{c}{$\uparrow$ Acc (\%)}
& \multicolumn{4}{c}{$\uparrow$ F1$_{\text{macro}}$ (\%)} \\
& \multicolumn{2}{c}{\textbf{Scratch}}
& \multicolumn{2}{c}{\textbf{MEUSLI}}
& \multicolumn{2}{c}{\textbf{Scratch}}
& \multicolumn{2}{c}{\textbf{MEUSLI}}
& \multicolumn{2}{c}{\textbf{Scratch}}
& \multicolumn{2}{c}{\textbf{MEUSLI}}
& \multicolumn{2}{c}{\textbf{Scratch}}
& \multicolumn{2}{c}{\textbf{MEUSLI}} \\
& \textbf{Mono} & \textbf{Multi}
& \textbf{Mono} & \textbf{Multi}
& \textbf{Mono} & \textbf{Multi}
& \textbf{Mono} & \textbf{Multi}
& \textbf{Mono} & \textbf{Multi}
& \textbf{Mono} & \textbf{Multi}
& \textbf{Mono} & \textbf{Multi}
& \textbf{Mono} & \textbf{Multi} \\
\hline
Italian
& 129.0 & 12.6 & 3.4 & 3.7
& 0.8 & 43.3 & 54.6 & 54.5
& 71.8 & 66.1 & 84.3 & 86.8
& 69.3 & 68.4 & 82.3 & 86.4 \\

Spanish
& 121.6 & 9.8 & 4.2 & 4.7
& 0.5 & 47.3 & 56.2 & 55.5
& 54.5 & 67.7 & 84.7 & 86.4
& 55.1 & 69.1 & 83.3 & 85.1 \\

Galician
& 158.4 & 34.0 & 10.2 & 11.1
& 0.9 & 34.3 & 46.8 & 47.5
& 54.3 & 67.0 & 83.6 & 86.1
& 42.3 & 69.0 & 81.6 & 85.0 \\

Czech
& 156.5 & 43.1 & 11.0 & 12.9
& 0.3 & 22.2 & 40.7 & 39.3
& 53.0 & 60.8 & 77.4 & 86.7
& 43.6 & 63.9 & 76.9 & 86.1 \\

Finnish
& 162.0 & 53.0 & 17.3 & 18.6
& 0.5 & 15.9 & 27.0 & 26.6
& 33.1 & 57.5 & 78.0 & 81.7
& 10.3 & 59.8 & 74.5 & 79.2 \\
\hline
\end{tabular}%
}
\end{table*}

Table~\ref{tab:meusli_multitask_scratch_vs_bootstrap} shows that monolingual multitask training from-scratch is insufficient under limited supervision, as evident from the poor ASR performance (e.g., 129.0\%--162.0\% WER across languages) and near-zero speech translation quality (0.3\%--0.9\% BLEU). In contrast, training a single multilingual model from scratch substantially improves performance through cross-lingual transfer, reducing WER to 9.8\%--53.0\% and increasing BLEU to 15.9\%--47.3\%. These results suggest that multilingual supervision can benefit speech-language alignment under data-scarce conditions.

Bootstrapping the pretrained MEUSLI projector provides further improvements across all three tasks and all five languages. Relative to multilingual training from-scratch, multilingual MEUSLI finetuning reduces ASR WER from 53.0\% to 18.6\% WER for Finnish, 43.1\% to 12.9\% for Czech, and 34.0\% to 11.1\% for Galician. Similarly, ST performance improves from 15.9\% to 26.6\% BLEU for Finnish, 22.2\% to 39.3\% BLEU for Czech, and 34.3\% to 47.5\% BLEU for Galician. TID accuracy also improves from 57.5\% to 81.7\% for Finnish, 60.8\% to 86.7\% for Czech, and 67.0\% to 86.1\% for Galician.

Both MEUSLI finetuning strategies (Mono vs Multi) can achieve strong performance across ASR, ST, and TID. This suggests that the multilingual representations learned during MEUSLI pretraining can be leveraged beyond speech recognition. 

In general, these findings suggest that MEUSLI can enable further adaptation to speech recognition, generation, and classification tasks even with less than five hours of task-specific data per language. Although language-specific finetuning appears to remain advantageous for ASR, the results indicate that multilingual finetuning can provide robust shared semantic representations for downstream understanding tasks.

\section{Limitations}
We note that the MEUSLI projector was originally pre-trained on multilingual ASR data from FLEURS (see Section \ref{ssec:exp_setup_datasets}), which overlaps with the SIB-Fleurs dataset used in this multitask extension. Although we adapt the projector to new downstream tasks rather than ASR alone, this pretraining data may partially contribute to the strong transfer performance observed and should be considered when interpreting the results.

\section{Conclusions}
In this work, we presented MEUSLI, a multilingual linear projector that bridges a pretrained speech encoder (Whisper) with pretrained multilingual LLMs, enabling end-to-end ASR across 28 European languages using fully open-source models. Experiments show that this approach extends prior English-only pipelines to multilingual settings with strong performance across high- and low-resource languages. 

We further showed that the multilingual projector serves as an effective initialization for fine-tuning on underrepresented languages and can bootstrap ASR capabilities for entirely new languages, even with extremely limited training data. In addition, we demonstrate that MEUSLI transfers effectively beyond ASR to speech translation and topic identification. Only a few hours of task-specific supervision per language are required for multilingual multitask adaptation.
%

Overall, these results highlight the scalability, robustness, and inclusivity of SLMs when paired with multilingual pretraining, providing a practical pathway towards LLM-based universal and open-source systems. 

\newpage
\section{Acknowledgments}
This work has received funding from the European Union’s Horizon Europe research and innovation program under the project ELOQUENCE (Grant Agreement No. 101135916). 
Views and opinions expressed are, however, those of the author(s) only and do not necessarily reflect those of the European Union or Research Executive Agency. Neither the European Union nor the granting authority can be held responsible for them.

\bibliographystyle{IEEEtran}
\bibliography{mybib}

\end{document}